# Variable transformations in consistent loss functions

Hristos Tyralis[1*], Georgia Papacharalampous[2]

[1]Support Command, Hellenic Air Force, Elefsina Air Base, 19 200, Elefsina, Greece (montchrister@gmail.com, hristos@itia.ntua.gr, https://orcid.org/0000-0002-8932-4997)

[2]Department of Land, Environment, Agriculture and Forestry, University of Padova, Viale dell'Università 16, 35020, Legnaro, Italy (papacharalampous.georgia@gmail.com, georgia.papacharalampous@unipd.it, https://orcid.org/0000-0001-5446-954X)

*Corresponding author

**Abstract**: The empirical use of variable transformations within (strictly) consistent loss functions is widespread, yet a theoretical understanding is lacking. To address this gap, we develop a theoretical framework that establishes formal characterizations of (strict) consistency for such transformed loss functions. Our analysis focuses on two interrelated cases: (a) transformations applied solely to the realization variable and (b) bijective transformations applied jointly to both the realization and prediction variables. These cases extend the well-established framework of transformations applied exclusively to the prediction variable, as formalized by Osband's revelation principle. We further develop analogous characterizations for (strict) identification functions. The resulting theoretical framework is broadly applicable to statistical and machine learning methodologies. For instance, we apply the framework to Bregman and expectile loss functions to interpret empirical findings from models trained with transformed loss functions and systematically construct new identifiable and elicitable functionals, which we term respectively $g$-transformed expectation and $g$-transformed expectile. Applications of the framework to simulated and real-world data illustrate its practical utility in diverse settings. By unifying theoretical insights with practical applications, this work advances principled methodologies for designing loss functions in complex predictive tasks.

**Keywords**: Bregman loss; consistent scoring function; elicitable functional; point forecasting; quantile loss.

## 1. Introduction

Predictive modeling is foundational to artificial intelligence, decision sciences, information sciences and operations research, where predictions inform resource allocation, system optimization, and risk mitigation strategies. The assessment of these predictions requires two distinct approaches, i.e., evaluating their statistical adequacy and comparing their performance. While identification functions [1] are used to assess adequacy, model comparison is typically performed using loss functions (also referred to as scoring or objective functions) for point predictions or scoring rules for probabilistic predictions. This work focuses specifically on identification and loss functions for point predictions. The use of these loss functions for performance comparison is well-established in statistical literature and machine learning ([2], [3], [4]).

Driven by domain-specific requirements, disciplines such as environmental science and technology have independently developed loss functions (often referred to as performance metrics or error metrics) tailored to their needs ([5], [6], [7]). These needs emphasize evaluating models' ability to predict phenomena such as extreme events or highly skewed variables. To address these challenges, loss functions of the form $S(g(z), g(y))$, where $S(z, y)$ evaluates a prediction $z$ against a realized outcome $y$, and $g$ is a bijection, are applied ([8], [9]). While empirical assessments of such transformations are well-documented ([8], [9]), their theoretical foundations remain unexplored, hindering broader implementation across disciplines.

Modelers typically train their models by minimizing average (realized) scores. From a statistical perspective, this training process can be interpreted as an application of $M$-estimation ([10], [11]) within a regression framework. Specifically, parameters are estimated by minimizing the average value of a loss function that quantifies the discrepancy between model predictions and observations.

Thus, loss functions serve a dual purpose: parameter estimation and prediction evaluation. [12] bridge these procedures, demonstrating that an $M$-estimator for a regression model is consistent if and only if the loss function used by the $M$-estimator is consistent for the target statistical functional (also referred to as statistical property). Common statistical functional examples include the mean, median, quantiles [13] and expectiles [14]. A loss function is (strictly) consistent for a statistical functional, if its expected value is (uniquely) minimized when the modeler follows the directive to predict the statistical functional [2]. The mean, median, quantile and expectile functionals are also identifiable, meaning that there exist (strict) identification functions whose expected value (uniquely) equals 0, when the modeler follows the directive to predict the functional ([2], [15]).

In predictive modeling, model comparisons are best achieved by disclosing the loss function a priori or specifying a statistical functional to predict for the dependent variable [2]. However, even when targeting a specific functional, different loss functions consistent for the functional can lead to different model rankings in practice [16]. Consequently, specifying both the desired functional and a consistent loss function for evaluation is critical [16].

Motivated by these considerations, we aim to bridge the gap between machine learning practices involving transformed predictions/realizations and statistical literature on consistent loss functions. We contribute in two ways:

a. Characterization of identification and consistent loss functions: Building on Osband's revelation principle [17] for transformed predictions and new theoretical results for transformed realizations (Theorems 3-4), we characterize identification/consistent loss functions for transformed predictions and realizations (Remark 1).

b. Specific functional families: We apply Theorems 3, 4 and Remark 1 to characterize families of (strict) identification and (strictly) consistent loss functions; including quantile and expectile cases (Section 4.2). These results yield new identifiable and elicitable functionals based on the squared-error and expectile loss function. They also explain empirical findings for transformed squared error loss functions ([8], [9]).

The remainder of the manuscript is structured as follows. To establish a conceptual framework, Section 2 reviews the key literature on loss functions. Section 3 reviews notation, identifiable/elicitable functionals, and strict identification/strictly consistent loss functions. Section 4 presents new theoretical results (Theorems 3, 4, Remark 1) and their application to specific identification/consistent loss function classes. Section 5 applies the theory to both simulated and real-world data, aiming to demonstrate its practical implications in real-world scenarios. Sections 6 and 7 discuss implications and conclude.

## 2. Literature review

This section summarizes the most relevant literature on loss functions, providing a foundation for the more detailed theoretical exposition in Section 3. The field of loss functions primarily concerns the evaluation of predictions, which can take the forms of point predictions or predictive distributions. These are evaluated using scoring functions and scoring rules, respectively. While both can be considered loss functions, this work focuses specifically on scoring functions for point predictions.

A key property a loss function can have is strict consistency (explained intuitively in Section 1 and formally in Section 3.2). The theory of consistent scoring functions has been extensively reviewed by [2]. An analogous property for scoring rules, known as propriety, has been thoroughly examined by [3], and is not discussed here. An earlier overview of both topics was provided by [18].

Although prediction evaluation and model estimation are distinct tasks, loss functions can serve both purposes. It has been established that when used for evaluation, the properties of consistency and propriety extend to model estimation. Specifically, estimators based on consistent loss functions and proper scoring rules are themselves consistent estimators (where "consistency" for estimators has a different definition than for evaluation). These equivalences have been explicitly demonstrated by [12] and [3] respectively.

The topic of transforming the prediction variable has been studied by [2] and [17], in the context of loss functions (see Section 3.5), while [19] examined the same problem for proper scoring rules. Empirical investigations of transformations, particularly those explored by [9] are extensive. This manuscript aims to provide a theoretical explanation for their findings.

Finally, Figure 1 illustrates the manuscript's workflow The theoretical development builds upon the strict consistency of a base loss function $S(z, y)$. From this foundation, Theorem 3 establishes the strict consistency of functions of the form $S(z, g(y))$. This result is then combined with Osband's revelation principle to prove Remark 1, which characterizes the consistency of loss functions of the form $S(g(z), g(y))$. Subsequent sections present theoretical and empirical applications derived from this framework.

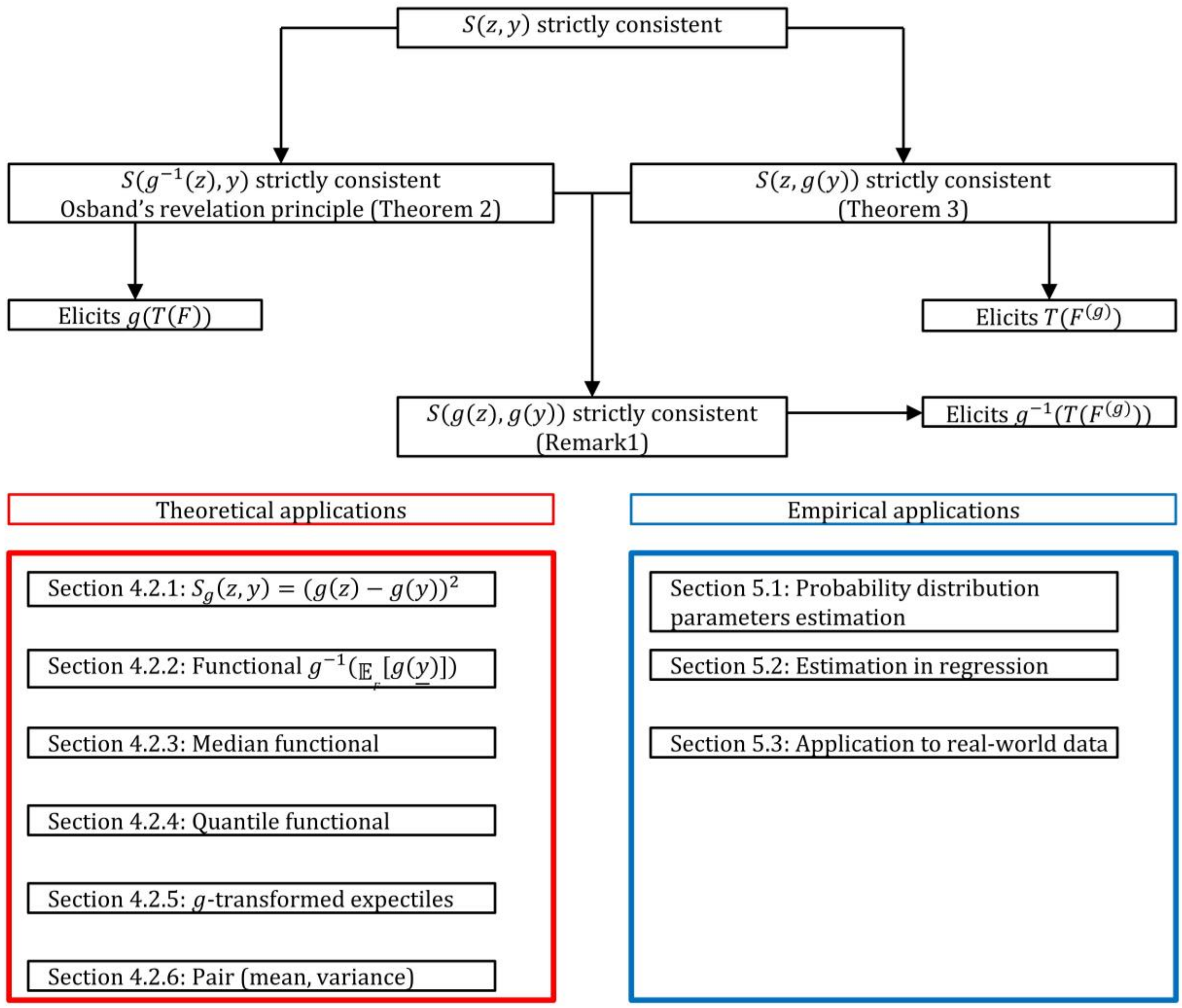

Figure 1. Workflow of the manuscript.

## 3. Notation and theory of consistent loss functions

This section outlines the theory of consistent loss functions and identification functions. Theoretical results are accompanied by practical examples. The discussion is limited to results essential for understanding the theoretical framework in Section 4, while we do not intend to provide a comprehensive overview of the topic. Appendix A introduces the notation and definitions for vector calculus used consistently throughout the manuscript.

### 3.1 Random variables and statistical functionals

Let $\underline{y}$ be a random variable. Hereinafter, random variables will be underlined. A materialization of $\underline{y}$ will be denoted by $y$. We write $\underline{y} \sim F$ to indicate that $\underline{y}$ has cumulative distribution function (CDF) $F$, i.e.

$$F(y) := P(\underline{y} \leq y) \tag{3.1}$$

A $k$-dimensional statistical functional (or simply a functional) $T$ [2] is a mapping

$$T: \mathcal{F} \to \mathcal{P}(D), F \mapsto T(F) \subseteq D \tag{3.2}$$

where $\mathcal{F}$ is a class of probability distributions. The functional maps each probability distribution $F \in \mathcal{F}$ to a subset $T(F)$ of the domain $D \subseteq \mathbb{R}^k$, that is an element of the power set $\mathcal{P}(D)$. $T(F)$ can be single-valued or set-valued.

**Running example 1.1**: An example of a single valued one-dimensional statistical functional is the mean of a probability distribution. It is defined by

$$E^{1/2}(F) := \mathbb{E}_F[\underline{y}], \tag{3.3}$$

where $\mathbb{E}_F[\cdot]$ is the expectation of a random variable $\underline{y}$ with CDF $F$. The mean functional is a special case of the expectile functional $E^\tau(F)$ for $\tau = 1/2$; see Section 3.8.2. Another example of a one-dimensional statistical functional is the $\tau$-quantile $Q^\tau(F)$ (or quantile at level $\tau$ [13]), which is a mapping:

$$Q^\tau: \mathcal{F} \to \mathcal{P}(D), D \subseteq \mathbb{R} \tag{3.4}$$

and is defined by:

$$Q^\tau(F) = \{z \in D: \lim_{y \uparrow z} F(y) \leq \tau \leq F(z)\}, D \subseteq \mathbb{R}, \tau \in (0,1), F \in \mathcal{F} \tag{3.5}$$

The $\tau$-quantile $Q^\tau(F)$ is generally a set-valued functional [20], since it is a closed bounded interval of $\mathbb{R}$ [21]. If $F$ is a strictly increasing function in $\mathbb{R}$, then $Q^\tau(F)$ is single-valued.■

## 3.2 Strictly consistent loss functions

Let $\boldsymbol{y} \in I \subseteq \mathbb{R}^d$ be a realization of a variable $\underline{\boldsymbol{y}}$ and $\boldsymbol{z} \in D \subseteq \mathbb{R}^k$ be a $k$-dimensional functional prediction. Predictions can be issued by either a time series model, a machine learning algorithm, or a physics-based model among others. Then a loss function $S$ is a mapping

$$S: D \times I \to \mathbb{R} \tag{3.6}$$

that assigns a penalty (loss) $S(\boldsymbol{z}, \boldsymbol{y})$ to a prediction $\boldsymbol{z}$ when $\boldsymbol{y}$ realizes. In the following, loss functions will be negatively oriented, i.e. the lower the penalty, the better the prediction. An appealing property of several loss functions is strict consistency.

**Definition 1** ([2], [18]): The loss function $S$ is $\mathcal{F}$-consistent for the functional $T$ if

$$\mathbb{E}_F[S(\boldsymbol{t}, \underline{\boldsymbol{y}})] \leq \mathbb{E}_F[S(\boldsymbol{z}, \underline{\boldsymbol{y}})] \ \forall \ F \in \mathcal{F}, \boldsymbol{t} \in T(F), \boldsymbol{z} \in D \tag{3.7}$$

It is strictly $\mathcal{F}$-consistent if it is $\mathcal{F}$-consistent and equality in eq. (3.7) implies that $\boldsymbol{z} \in T(F)$.■

The definition states that if a modeler receives a directive to predict a functional, and uses an $\mathcal{F}$-consistent loss function to evaluate its prediction, then the loss function is minimized in expectation when the prediction is optimal (i.e. equal to the value of the functional). If the loss function is strictly $\mathcal{F}$-consistent, then it is uniquely minimized in expectation at the optimal prediction.

The following theorem demonstrates the relationship between finding optimal predictions and evaluating predictions. It states that the classes of consistent loss functions for a specified functional and of loss functions under which the functional is an optimal prediction are identical [2].

**Theorem 1** ([2]): The loss function $S$ is $\mathcal{F}$-consistent for the functional $T$ if and only if, given any $F \in \mathcal{F}$, any $\boldsymbol{z} \in T(F)$ is an optimal point prediction under $S$.■

The following definition of elicitability of a functional explains the correspondence between specific functionals and loss functions.

**Definition 2** ([2], [17], [22]): The functional $T$ is elicitable relative to the class $\mathcal{F}$ if there exists a loss function $S$ that is strictly $\mathcal{F}$-consistent for $T$.■

For some of the following results, the assumption that the loss function $S(\boldsymbol{z}, \boldsymbol{y})$ is bounded below is required [2]. We can assume nonnegativity of the loss function $S$ without loss of generality, since properties are transferrable to any lower bounded function. The properties of a consistent loss function do not change if multiplied with a positive constant and a function of $\boldsymbol{y}$ is added [2].

Loss functions with the property of homogeneity are appealing.

**Definition 3** ([2]): A loss function $S$ is homogeneous of order $b$ if

$$S(c\boldsymbol{z}, c\boldsymbol{y}) = |c|^b S(\boldsymbol{z}, \boldsymbol{y}) \; \forall \; \boldsymbol{z} \in D, \boldsymbol{y} \in I, c \in \mathbb{R} \quad (3.8)$$

■

Homogeneity order describes a function's scaling behavior. For example, functions of order 0 ignore scale and measurement units, of order 1 scale linearly, and of order 2 penalize large discrepancies more [23].

**Running example 1.2**: The mean is an elicitable functional for the class $\mathcal{F}$ of probability distributions with finite second moment. A strictly $\mathcal{F}$-consistent loss function for the mean functional is the squared error loss function

$$S_{\mathrm{SE}}(z, y) := (z - y)^2 \quad (3.9)$$

A simple proof is given below:

$$\mathbb{E}_F[S_{\text{SE}}(z,\underline{y})] - \mathbb{E}_F[S_{\text{SE}}(\mathbb{E}_F[\underline{y}],\underline{y})] = (z^2 - 2z\mathbb{E}_F[\underline{y}] + \mathbb{E}_F^2[\underline{y}] - \mathbb{E}_F^2[\underline{y}] + 2\mathbb{E}_F^2[\underline{y}] - \mathbb{E}_F^2[\underline{y}]) \Rightarrow \tag{3.10}$$

$$\mathbb{E}_F[S_{\text{SE}}(z,\underline{y})] - \mathbb{E}_F[S_{\text{SE}}(\mathbb{E}_F[\underline{y}],\underline{y})] = (z - \mathbb{E}_F[\underline{y}])^2 \tag{3.11}$$

Therefore,

$$\mathbb{E}_F[S_{\text{SE}}(\mathbb{E}_F[\underline{y}],\underline{y})] \leq \mathbb{E}_F[S_{\text{SE}}(z,\underline{y})]\ \forall\ F \in \mathcal{F}, z \in \mathbb{R} \tag{3.12}$$

and

$$\mathbb{E}_F[S_{\text{SE}}(z,\underline{y})] = \mathbb{E}_F[S_{\text{SE}}(\mathbb{E}_F[\underline{y}],\underline{y})] \Leftrightarrow z = \mathbb{E}_F[\underline{y}] \tag{3.13}$$

that completes the proof.

The median functional, i.e. the 1/2-quantile $Q^{1/2}(F)$ is also elicitable relative to the class $\mathcal{F}$ of probability distributions with finite first moment [2]. A strictly $\mathcal{F}$-consistent loss function for the median functional is the absolute error loss function

$$S_{\text{AE}}(z,y) := |z - y| \tag{3.14}$$

■

## 3.3 Identification functions

We assume that there exists a function ([2], [15]):

$$V: D \times I \to \mathbb{R} \tag{3.15}$$

**Definition 4** ([2], [15]): $V$ is said to be an $\mathcal{F}$-identification function for the functional $T(F)$ if

$$\mathbb{E}_F[V(T(F),\underline{\boldsymbol{y}})] = 0 \tag{3.16}$$

It is a strict $\mathcal{F}$-identification function for the functional $T(F)$ if

$$\mathbb{E}_F[V(\boldsymbol{z},\underline{\boldsymbol{y}})] = 0 \Leftrightarrow \boldsymbol{z} \in T(F) \tag{3.17}$$

If there is a strict $\mathcal{F}$-identification function for the functional $T(F)$, then $T$ is identifiable.■

The importance of the identifiability property of a functional lies in the ability to assess the reliability of identifiable functionals' predictions in absolute terms [20], also known as calibration in the statistical literature ([23], [24]). Loss functions, on the other hand, are used for prediction comparison and ranking [15]. Connections between identifiable and elicitable functionals (explained later in Section 3.4) will be explored, as they allow characterizing properties of loss functions.

**Running example 1.3**: The mean is an identifiable functional for the class $\mathcal{F}$ of probability distributions with finite second moment. A strict $\mathcal{F}$-identification function for the mean functional is ([2])

$$V_{E^{1/2}}(z, y) := z - y \tag{3.18}$$

A simple proof is given below:

$$\mathbb{E}_F[V_{E^{1/2}}(z, \underline{y})] = 0 \Leftrightarrow z = \mathbb{E}_F[\underline{y}] \tag{3.19}$$

An interpretation of the identifiability property of the mean functional is that the expectation of the identification function $V_{E^{1/2}}$ becomes equal to 0 when a prediction is optimal, i.e. equal to the mean functional.

The median is an identifiable functional for the class $\mathcal{F}$ of probability distributions with finite first moment, with strict $\mathcal{F}$-identification function ([2])

$$V_{Q^{1/2}}(z, y) = \mathbb{1}\{z \geq y\} - 1/2 \tag{3.20}$$

where $\mathbb{1}(\cdot)$ denotes the indicator function.■

## 3.4 Osband's principle

A strictly $\mathcal{F}$-consistent loss function $S$ and a strict $\mathcal{F}$-identification function $V$ are linked through ([2], [15], [17]):

$$\frac{\partial S(z,y)}{\partial z} = h(z)V(z, y) \tag{3.21}$$

where $h$ is a nonnegative function and $V$ is an oriented function, defined as follows:

**Definition 5** ([15], [25]): The function $V$ is an oriented strict $\mathcal{F}$-identification function if it is a strict $\mathcal{F}$-identification function for the functional $T(F)$ and, moreover $\mathbb{E}_F[V(z, \underline{y})] > 0$ if and only if $z > T(F)$ $\forall$ $F \in \mathcal{F}$, $z \in D \subseteq \mathbb{R}$.■

Osband [17] introduced this characterization, now known as Osband's principle. Osband's principle importance lies in the fact that it allows constructing strictly $\mathcal{F}$-consistent loss functions from strict $\mathcal{F}$-identification functions for a functional $T(F)$ ([15], [25]). For the case of $k = 1$ (i.e. of the one-dimensional functional), if $V$ is a strict $\mathcal{F}$-identification function, then either $V$ or $-V$ is oriented. Then specifying a strictly positive function $h$ and integrating the right part of eq. (3.21) yields a strictly $\mathcal{F}$-consistent loss function.

**Running example 1.4**: The strictly consistent $S_{\mathrm{SE}}$ loss function and the strict identification function $V_{E^{1/2}}$ of the mean functional are connected by:

$$\frac{\partial S_{\mathrm{SE}}(z,y)}{\partial z} = 2V_{E^{1/2}}(z,y) \tag{3.22}$$

■

## 3.5 Osband's revelation principle

Strictly consistent loss functions and strict identification functions can be constructed by applying bijective transformations to the prediction variable $\boldsymbol{z}$ of existing strictly consistent loss functions and identification functions. Similar constructions apply to elicitable and identifiable functionals. This principle, known as Osband's revelation principle [17], is summarized as follows:

**Theorem 2** ([2], [17], [26], [27]): Let $g: D \to D'$, $D, D' \subseteq \mathbb{R}^k$ be a bijection with inverse $g^{-1}: D' \to D$. Let the functional $T$ be defined on a class $\mathcal{F}$ of probability distributions with CDF $F$ and the transformation $T_g$ of $T$ be defined by $T_g: \mathcal{F} \to D', F \mapsto T_g(F) = g(T(F))$. Then the following holds:

a. $T$ is elicitable if and only if $T_g$ is elicitable.

b. $T$ is identifiable if and only if $T_g$ is identifiable.

c. The loss function $S: D \times I \to \mathbb{R}, (\boldsymbol{z}, \boldsymbol{y}) \mapsto S(\boldsymbol{z}, \boldsymbol{y}), I \subseteq \mathbb{R}^d$ is $\mathcal{F}$-consistent for $T$, if and only if the loss function $S_g: D' \times I \to \mathbb{R}, (\boldsymbol{z}, \boldsymbol{y}) \mapsto S_g(\boldsymbol{z}, \boldsymbol{y}) = S(g^{-1}(\boldsymbol{z}), \boldsymbol{y})$ is $\mathcal{F}$-consistent for $T_g$.

d. The loss function $S: D \times I \to \mathbb{R}, (\boldsymbol{z}, \boldsymbol{y}) \mapsto S(\boldsymbol{z}, \boldsymbol{y}), I \subseteq \mathbb{R}^d$ is strictly $\mathcal{F}$-consistent for $T$, if and only if the loss function $S_g: D' \times I \to \mathbb{R}, (\boldsymbol{z}, \boldsymbol{y}) \mapsto S_g(\boldsymbol{z}, \boldsymbol{y}) = S(g^{-1}(\boldsymbol{z}), \boldsymbol{y})$ is strictly $\mathcal{F}$-consistent for $T_g$.

e. The function $V: D \times I \to \mathbb{R}^k, (\boldsymbol{z}, \boldsymbol{y}) \mapsto V(\boldsymbol{z}, \boldsymbol{y}), I \subseteq \mathbb{R}^d$ is a $\mathcal{F}$-identification function for $T$, if and only if the function $V_g: D' \times I \to \mathbb{R}^k, (\boldsymbol{z}, \boldsymbol{y}) \mapsto V_g(\boldsymbol{z}, \boldsymbol{y}) = V(g^{-1}(\boldsymbol{z}), \boldsymbol{y})$ is a $\mathcal{F}$-identification function for $T_g$.

f. The function $V: D \times I \to \mathbb{R}^k, (\boldsymbol{z}, \boldsymbol{y}) \mapsto V(\boldsymbol{z}, \boldsymbol{y}), I \subseteq \mathbb{R}^d$ is a strict $\mathcal{F}$-identification function for $T$, if and only if the function $V_g: D' \times I \to \mathbb{R}^k, (\boldsymbol{z}, \boldsymbol{y}) \mapsto V_g(\boldsymbol{z}, \boldsymbol{y}) = V(g^{-1}(\boldsymbol{z}), \boldsymbol{y})$ is a strict $\mathcal{F}$-identification function for $T_g$.■

**Running example 1.5**: Let $g: \mathbb{R} \to \mathbb{R}^+, t \mapsto g(t) = \exp(t)$. Then the inverse function $g^{-1}$ is defined by $g^{-1}: \mathbb{R}^+ \to \mathbb{R}, t \mapsto g^{-1}(t) = \log(t)$.

Let the functional $T: \mathcal{F} \to \mathbb{R}, F \mapsto T(F) = \mathbb{E}_F[\underline{y}] \subseteq \mathbb{R}$ be defined on the class $\mathcal{F}$ of probability distributions with positive support and finite second moment and the transformation $T_g$ of $T$ be defined by $T_g\colon \mathcal{F} \to \mathbb{R}^+, F \mapsto T_g(F) = \exp(\mathbb{E}_F[\underline{y}]) \subseteq \mathbb{R}^+$.

The loss function $S_{\mathrm{SE}}\colon \mathbb{R} \times \mathbb{R} \to [0, \infty), (z, y) \mapsto S_{\mathrm{SE}}(z, y) = (z - y)^2$ is strictly $\mathcal{F}$-consistent for the functional $T$; therefore, the loss function $S_g\colon \mathbb{R}^+ \times \mathbb{R} \to [0, \infty), (z, y) \mapsto S_g(z, y) = (\log(z) - y)^2$ is strictly $\mathcal{F}$-consistent for the functional $T_g$.■

### 3.6 Realized (average) and skill scores

In practical situations, when one has to compare two predictions of a functional, she/he is given a test set with realizations $\boldsymbol{y} = (y_1, \dots, y_n)^{\mathrm{T}}$ of the random variable $\underline{y}$. Assuming that the respective predictions of the functional are $\boldsymbol{z} = (z_1, \dots, z_n)^{\mathrm{T}}$, a summary measure of the predictive performance takes the form of the (negatively oriented) realized (average) score $\bar{S}$ [2]:

$$\bar{S}(\boldsymbol{z}, \boldsymbol{y}) = (1/n) \textstyle\sum_{i=1}^{n} S(z_i, y_i) \tag{3.23}$$

When choosing between two competing predictions, the one with the lowest average score is preferable.

When predictions from multiple methods have to be compared, skill scores can be used for evaluation. A skill score takes the form:

$$\bar{S}_{\mathrm{skill}}(\boldsymbol{z}, \boldsymbol{y}; \mathrm{ref}) := (\bar{S}(\boldsymbol{z}, \boldsymbol{y}) - \bar{S}(\boldsymbol{z}_{\mathrm{ref}}, \boldsymbol{y}))/(\bar{S}(\boldsymbol{z}_{\mathrm{optimal}}, \boldsymbol{y}) - \bar{S}(\boldsymbol{z}_{\mathrm{ref}}, \boldsymbol{y})) \tag{3.24}$$

where $\boldsymbol{z}_{\mathrm{ref}}$ are predictions of a reference method (frequently the simpler method among the competing ones) and $\boldsymbol{z}$ are predictions of the method of interest. If the loss function $S$ is minimized to 0, then for an optimal prediction $\boldsymbol{z}_{\mathrm{optimal}}$ we have $\bar{S}(\boldsymbol{z}_{\mathrm{optimal}}, \boldsymbol{y}) = 0$, hence the skill score takes the form:

$$\bar{S}_{\mathrm{skill}}(\boldsymbol{z}, \boldsymbol{y}; \mathrm{ref}) := 1 - \bar{S}(\boldsymbol{z}, \boldsymbol{y})/\bar{S}(\boldsymbol{z}_{\mathrm{ref}}, \boldsymbol{y}) \tag{3.25}$$

A positive skill score suggests that the proposed method outperforms the reference method, while a negative score indicates the opposite. The skill score attains its maximum value equal to 1 at a perfect prediction $\boldsymbol{z} = \boldsymbol{y}$. Skill scores maintain prediction rankings, while the notions of consistency and elicitability remain valid, at least for large test samples [2]. Quantitatively, skill scores measure the performance improvement of one prediction relative to another, as a percentage of the potential improvement from a perfect prediction [28].

**Running example 1.6**: In the case of the mean functional, specifying the squared error loss function to evaluate predictions and using mean climatology on the test set as a reference method in skill score's eq. (3.25) yields the Nash-Sutcliffe efficiency (NSE), introduced by [29]:

$$\mathrm{NSE}(\boldsymbol{z}, \boldsymbol{y}) := 1 - \bar{S}_{\mathrm{SE}}(\boldsymbol{z}, \boldsymbol{y}) / \bar{S}_{\mathrm{SE}}(\boldsymbol{1}\bar{\boldsymbol{y}}, \boldsymbol{y}) \tag{3.26}$$

The NSE is a skill score; therefore, it retains properties discussed earlier.■

## 3.7 $M$-estimation

Previous sections focused on evaluating point predictions with loss functions. However, they did not address how to use loss functions for parameter estimation in regression settings, and what the effect of parameter estimation using specific loss functions is on predictions of the regression algorithm. Properties of consistent loss functions and consistent estimators have been linked by [12]. We examine regression settings which model the effect of predictors $\underline{\boldsymbol{x}}$ on a response variable $\underline{y}$. For a correctly specified model, we might be interested in estimating the conditional probability distribution of $\underline{y}$, namely $F_{\underline{y}|\underline{\boldsymbol{x}}}$ in a distributional regression setting [30]. Nevertheless, we might also be interested in a specific functional $T(F_{\underline{y}|\underline{\boldsymbol{x}}})$ of the conditional distribution. For instance, predictive quantiles at multiple levels can substitute the conditional probability distribution.

We assume that a correctly specified semiparametric model $m(\underline{\boldsymbol{x}}, \boldsymbol{\theta})$ satisfies for some unique parameter $\boldsymbol{\theta}_0 \in \boldsymbol{\Theta}$

$$T(F_{\underline{y}|\underline{\boldsymbol{x}}}) = m(\underline{\boldsymbol{x}}, \boldsymbol{\theta}_0) \tag{3.27}$$

Model $m(\underline{\boldsymbol{x}}, \boldsymbol{\theta})$ is called semiparametric because the form of the conditional probability distribution need not to be specified, although it may belong to a potentially wide family $\mathcal{F}$. Consequently, parameterization is restricted to the model's form itself, e.g. linear models or decision trees [12]. Semiparametric models offer more flexibility compared to parametric models because their estimation procedures are common for wide classes of probability distributions, for which the functional exists. This prevents modeling restrictions to potentially misspecified distributions. However, that comes at the cost of reduced ability to extrapolate [4].

In practical situations, one specifies the general form of the model $m(\underline{\boldsymbol{x}}, \boldsymbol{\theta})$ but the parameter $\boldsymbol{\theta}$ still needs to be estimated from available data. Assuming that we have $l$ realizations of the vector $\underline{\boldsymbol{x}}$ and the dependent variable $\underline{y}$, an $M$-estimator of $\boldsymbol{\theta}_0$ is ([10], [11], [12], [31])

$$\widehat{\boldsymbol{\theta}}_l = \underset{\boldsymbol{\theta} \in \boldsymbol{\Theta}}{\arg\min} (1/l) \sum_{i=1}^{l} S(m(\boldsymbol{x}_i, \boldsymbol{\theta}), y_i) \tag{3.28}$$

where $S$ is a loss function. A substantive condition on $S$, for consistency of $\widehat{\boldsymbol{\theta}}_l$ is that ([11, p. 2123], [12])

$$\mathbb{E}_F[S(m(\underline{\boldsymbol{x}}, \boldsymbol{\theta}_0), \underline{y})] \leq \mathbb{E}_F[(m(\underline{\boldsymbol{x}}, \boldsymbol{\theta}), \underline{y})] \; \forall \underline{\boldsymbol{x}}, \underline{y} \in I, \boldsymbol{\theta} \in \boldsymbol{\Theta} \tag{3.29}$$

which is called model consistency of $S$ for the model $m$. Strict equality in eq. (3.29) for $\boldsymbol{\theta} = \boldsymbol{\theta}_0$ implies strict model consistency of $S$ for the model $m$.

A result from [12] establishes that a loss function $S$ is (strictly) model consistent for the model $m$, if and only if it is (strictly) consistent for the target functional $T$. That result has some important implications, for our subsequent analysis as it connects the procedures of estimation and prediction evaluation. If one aims to predict a functional with a semiparametric model, then she/he has to estimate the parameters using an $M$-estimator with a consistent loss function for the functional.

**Running example 1.7**: A consistent $M$-estimator for the parameter $\boldsymbol{\theta}_0$ of a distribution $F$ is

$$\widehat{\boldsymbol{\theta}}_l = \underset{\boldsymbol{\theta} \in \boldsymbol{\Theta}}{\arg\min} (1/l) \sum_{i=1}^{l} S(\boldsymbol{\theta}, y_i) \tag{3.30}$$

Eq. (3.30) can be viewed as a special case of a semiparametric model, where the model is equal to a constant parameter. If $S$ is a strictly $\mathcal{F}$-consistent loss function for the functional $T(F)$, then $\widehat{\boldsymbol{\theta}}_l$ is a strictly consistent estimator for the parameter $T(F)$ of the probability distribution $F$. For the case of the $S_{\mathrm{SE}}$ loss function, the well-known estimator of the mean of a probability distribution arises from eq. (3.30).

$$\widehat{\boldsymbol{\theta}}_l = \underset{\boldsymbol{\theta} \in \boldsymbol{\Theta}}{\arg\min} (1/l) \sum_{i=1}^{l} S_{\mathrm{SE}}(\boldsymbol{\theta}, y_i) \tag{3.31}$$

For the class of symmetric probability distributions, the mean and median functionals are identical. Therefore $\widehat{\boldsymbol{\theta}}_l$ is also a strictly consistent estimator of the median for this class ([12], [32]).■

## 3.8 Quantiles, expectiles and respective consistent loss functions

So far, we have reviewed examples of elicitable functionals, strictly consistent loss functions, and strict identification functions. However, the class of consistent loss functions for a given functional can include multiple distinct members. Below, we present several representative cases commonly encountered in practice. This list is not intended to be exhaustive.

*3.8.1 Quantiles*

The quantile functional has been defined in eq. (3.5). The general class of consistent loss functions for the quantile functional $Q^\tau(F)$ is defined by ([2], [33], [34]):

$$S_Q(z, y; \tau, g) := (\mathbb{1}(z \geq y) - \tau)(g(z) - g(y)) \quad (3.32)$$

where $g$ is a nondecreasing function. The generalized piecewise linear (GPL) loss function of order $\tau \in (0,1)$ $S_Q$ is strictly $\mathcal{F}$-consistent for the quantile $Q^\tau(F)$, if $g$ is strictly increasing and $\mathcal{F}$ is the family of probability distributions with finite first moment. The corresponding strict $\mathcal{F}$-identification function is defined by [2]:

$$V_{Q^\tau}(z, y) := \mathbb{1}(z \geq y) - \tau \quad (3.33)$$

$S_Q(z, y; \tau, t)$ is the asymmetric piecewise linear loss function (also termed as tick-loss or quantile loss in the literature) which is strictly consistent for the quantile $Q^\tau(F)$ [35], and lies at the heart quantile regression [13]:

$$S_Q(z, y; \tau, t) = (\mathbb{1}\{z \geq y\} - \tau)(z - y) \quad (3.34)$$

The absolute error loss function also arises from the GPL loss function:

$$S_{\mathrm{AE}}(z, y) = S_Q(z, y; 1/2, 2t) \quad (3.35)$$

*3.8.2 Expectiles*

The $\tau$-expectile $E^\tau(F)$ ([14], [21], [36], [37]) is defined by:

$$E^\tau(F) := \{z \in I : \tau \int_z^\infty (y - z)\mathrm{d}F(y) = (1 - \tau) \int_{-\infty}^z (z - y)\mathrm{d}F(y)\} \quad (3.36)$$

or equivalently

$$E^\tau(F) = \{z \in I : \tau\mathbb{E}_F[\kappa_{0,\infty}(\underline{y} - z)] = (1 - \tau)\mathbb{E}_F[\kappa_{0,\infty}(z - \underline{y})\} \quad (3.37)$$

where $\kappa_{a,b}(t)$ is the capping function, defined by:

$$\kappa_{a,b}(t) := \max\{\min\{t, b\}, -a\} \forall t \in \mathbb{R}, a, b \in [0, \infty] \quad (3.38)$$

or equivalently

$$\kappa_{a,b}(t) = \begin{cases} -a, t \leq -a \\ t, -a < t \leq b \\ b, t > b \end{cases} \quad (3.39)$$

The general class of consistent loss functions for the expectile functional $E^\tau(F)$ is defined by [2]:

$$S_E(z, y; \tau, \varphi) := |\mathbb{1}(z \geq y) - \tau|(\varphi(y) - \varphi(z) + \varphi'(z)(\mathrm{z} - \mathrm{y})) \quad (3.40)$$

where $\varphi$ is a convex function with subgradient $\varphi'$. $S_E$ is strictly $\mathcal{F}$-consistent for the expectile $E^{\tau}(F)$, if $\varphi$ is strictly convex, and $\mathcal{F}$ is the family of probability distributions with finite second moment. The corresponding strict $\mathcal{F}$-identification function is defined by [2]:

$$V_{E^{\tau}}(z, y) := 2|\mathbb{1}(z \geq y) - \tau|(z - y) \tag{3.41}$$

The asymmetric piecewise quadratic loss function $S_E(z, y; \tau, t^2)$ is strictly consistent for the expectile $E^{\tau}(F)$ and lies at the heart of expectile regression [14]:

$$S_E(z, y; \tau, t^2) = |\mathbb{1}\{z \geq y\} - \tau|(z - y)^2 \tag{3.42}$$

The squared error loss function is a special case of $S_E(z, y; \tau, \varphi)$ up to a multiplicative constant:

$$S_{\mathrm{SE}}(z, y) = 2S_E(z, y; 1/2, t^2) \tag{3.43}$$

The class of loss functions

$$S_E(z, y; 1/2, \varphi) = (1/2)(\varphi(y) - \varphi(z) + \varphi'(z)(z - y)) \tag{3.44}$$

that arises from eq. (3.40) for $\tau = 1/2$, is consistent for the mean functional [38]. Functions of the form $2S_E(z, y; 1/2, \varphi)$ are referred as Bregman functions by [39].

*3.8.3 The pair (mean, variance)*

Quantile and expectile loss functions apply to one-dimensional functionals and one-dimensional variables. Here, we consider an example of a two-dimensional functional: the pair $(\mathbb{E}_F[\underline{y}], \mathrm{Var}_F[\underline{y}])$ representing the mean $\mathbb{E}_F[\underline{y}]$ and variance $\mathrm{Var}_F[\underline{y}]$. This pair is both identifiable and elicitable. Specifically, it is identified by the mean-variance strict identification function ([2], [32], [40])

$$V_{\mathrm{mv}}(x_1, x_2, y) := (x_1 - y, x_2 + x_1^2 - y^2) \tag{3.45}$$

and elicited by the mean-variance strictly consistent loss function ([2], [17], [32])

$$S_{\mathrm{mv}}(x_1, x_2, y) := x_2^{-2}(x_1^2 - 2x_2 - 2x_1 y + y^2) \tag{3.46}$$

where $(x_1, x_2)$ denotes predictions for the pair $(\mathbb{E}_F[\underline{y}], \mathrm{Var}_F[\underline{y}])$.

## 4. Transforming realizations/predictions in loss/identification functions

The empirical exploration of transformations applied to both the predictor and realization variables in loss and identification functions has been extensively studied. However, extant theory has focused primarily on transformations of the prediction variables, guided by Osband's revelation principle [17]. In Section 4.1, we extend this framework to include transformations of the realization variable alone and simultaneous transformations of both independent variables. Section 4.2 then illustrates the broad spectrum of practical applications enabled by these theoretical extensions.

### 4.1 Characterizations of loss and identification functions

The next theorem characterizes loss functions that arise from applying a transformation to their realization variables. This is a general result, enabling the characterization of loss functions where the realization variable itself is transformed. For the proof, see Appendix B.

**Theorem 3**: Let the functional $T: \mathcal{F} \to \mathcal{P}(D), F \mapsto T(F) \subseteq D, D \subseteq \mathbb{R}^k$ of a random variable $\underline{\boldsymbol{y}}$ be defined on a class $\mathcal{F}$ of probability distributions with CDF $F$. Consider a measurable function $g: I' \to I'', I', I'' \subseteq \mathbb{R}^d$. Let $\mathcal{F}^{(g)} \subseteq \mathcal{F}$ denote the subclass of the probability distributions in $\mathcal{F}$ which are such that $\underline{\boldsymbol{y}}^{(g)} = g(\underline{\boldsymbol{y}})$ has CDF $F^{(g)}$ and $F^{(g)} \in \mathcal{F}$. Define the functional $T^{(g)}: \mathcal{F}^{(g)} \to \mathcal{P}(D), F^{(g)} \mapsto T^{(g)}(F^{(g)}) = T(F^{(g)})$ on this subclass $\mathcal{F}^{(g)}$. Then the following holds:

a. If $T$ is elicitable, then $T^{(g)}$ is elicitable.

b. If $S: D \times I'' \to \mathbb{R}, (\boldsymbol{z}, \boldsymbol{y}) \mapsto S(\boldsymbol{z}, \boldsymbol{y})$ is $\mathcal{F}$-consistent for $T$, then $S^{(g)}: D \times I' \to \mathbb{R}, (\boldsymbol{z}, \boldsymbol{y}) \mapsto S^{(g)}(\boldsymbol{z}, \boldsymbol{y}) = S\big(\boldsymbol{z}, g(\boldsymbol{y})\big)$ is $\mathcal{F}^{(g)}$-consistent for $T^{(g)}$.

c. If $S$ is strictly $\mathcal{F}$-consistent for $T$, then $S^{(g)}$ is strictly $\mathcal{F}^{(g)}$-consistent for $T^{(g)}$.

d. If $g$ is a bijection with inverse $g^{-1}: I'' \to I'$, there is a bijection between classes $\mathcal{F}$ and $\mathcal{F}^{(g)}$, while the implications in parts (a), (b) and (c) become equivalences.■

The next example demonstrates an application of Theorem 3 for the case where the log transformation is applied to the realization variable of the loss function. It illustrates how a new strictly consistent loss function arises from the squared error loss function and how a new elicitable functional is constructed from the mean functional.

**Running example 2.1**: Let $g: \mathbb{R}^+ \to \mathbb{R}, t \mapsto g(t) = \log(t)$. The function $g$ is a bijection with inverse $g^{-1}: \mathbb{R} \to \mathbb{R}^+, t \mapsto g(t) = \exp(t)$. Let the functional $T: \mathcal{F} \to \mathbb{R}, F \mapsto T(F) = \mathbb{E}_F[\underline{y}] \subseteq \mathbb{R}$ be defined on the class $\mathcal{F}$ of probability distributions with finite second moment. Let $\mathcal{F}^{(g)} \subseteq \mathcal{F}$ denote the class of the probability distributions in $\mathcal{F}$ which are such that $\underline{y}^{(g)} = \log(\underline{y})$ has CDF $F^{(g)} \in \mathcal{F}^{(g)}$. Let the transformation $T^{(g)}$ of $T$ be defined by $T^{(g)}: \mathcal{F}^{(g)} \to \mathbb{R}, F \mapsto T^{(g)}(F^{(g)}) = \mathbb{E}_F[\log(\underline{y})] \subseteq \mathbb{R}$, because $T^{(g)}(F^{(g)}) = \mathbb{E}_{F^{(g)}}[\underline{y}^{(g)}] = \mathbb{E}_F[\log(\underline{y})]$ due to the law of the unconscious statistician (LOTUS).

The loss function $S_{\mathrm{SE}}: \mathbb{R} \times \mathbb{R} \to [0, \infty), (z, y) \mapsto S_{\mathrm{SE}}(z, y) = (z - y)^2$ is strictly $\mathcal{F}$-consistent for the functional $T$; therefore, the loss function $S_g: \mathbb{R} \times \mathbb{R}^+ \to [0, \infty), (z, y) \mapsto S_g(z, y) = (z - \log(y))^2$ (the support of $z$ remains $\mathbb{R}$, since it is not transformed) is strictly $\mathcal{F}^{(g)}$-consistent for the functional $T^{(g)}$.

Moreover, $\mathcal{F}^{(g)}$ is the class of probability distributions $F^{(g)}$ of $\underline{y}^{(g)}$, with finite expectation $\mathbb{E}_{F^{(g)}}[\underline{y}^{(g)}]$ or equivalently finite expectation $\mathbb{E}_F[\log(\underline{y})]$.■

The next theorem complements Theorem 3 by addressing the case of identification functions where a transformation is applied to their realization variable. For the proof, see Appendix B.

**Theorem 4**: Let the functional $T: \mathcal{F} \to \mathcal{P}(D), F \mapsto T(F) \subseteq D, D \subseteq \mathbb{R}^k$ of a random variable $\underline{\boldsymbol{y}}$ be defined on a class $\mathcal{F}$ of probability distributions with CDF $F$. Consider a function $g: I' \to I'', I', I'' \subseteq \mathbb{R}^d$. Let $\mathcal{F}^{(g)} \subseteq \mathcal{F}$ denote the subclass of the probability distributions in $\mathcal{F}$ which are such that $\underline{\boldsymbol{y}}^{(g)} = g(\underline{\boldsymbol{y}})$ has CDF $F^{(g)}$ and $F^{(g)} \in \mathcal{F}$. Define the functional $T^{(g)}: \mathcal{F}^{(g)} \to \mathcal{P}(D), F^{(g)} \mapsto T^{(g)}(F^{(g)}) = T(F^{(g)})$ on this subclass $\mathcal{F}^{(g)}$. Then the following holds:

a. If $T$ is identifiable, then $T^{(g)}$ is identifiable.

b. If $V: D \times I'' \to \mathbb{R}, (\boldsymbol{z}, \boldsymbol{y}) \mapsto V(\boldsymbol{z}, \boldsymbol{y})$ is an $\mathcal{F}$-identification function for $T$, then $V^{(g)}: D \times I' \to \mathbb{R}, (\boldsymbol{z}, \boldsymbol{y}) \mapsto V^{(g)}(\boldsymbol{z}, \boldsymbol{y}) = V\big(\boldsymbol{z}, g(\boldsymbol{y})\big)$ is an $\mathcal{F}^{(g)}$-identification function for $T^{(g)}$.

c. If $V$ is a strict $\mathcal{F}$-identification function for $T$, then $V^{(g)}$ is a strict $\mathcal{F}^{(g)}$-identification function for $T^{(g)}$.

d. If $g$ is a bijection with inverse $g^{-1}: I'' \to I'$, there is a bijection between classes $\mathcal{F}$ and $\mathcal{F}^{(g)}$, while the implications in parts (a), (b) and (c) become equivalences.■

Applying Theorem 4 enables the construction of new strict identification functions and identifiable functionals using a process analogous to that illustrated in Running Example 2.1.

The following remark presents the central result of this manuscript. It enables the characterization of consistent loss functions, identification functions, and their corresponding elicitable and identifiable functionals, which are generated by bijective transformations of the prediction and realization variables within associated consistent loss functions or identification functions. This result follows directly from Osband's revelation principle and Theorems 3 and 4; therefore, its proof is omitted.

**Remark 1**: Let $g: D \to D'$, $D, D' \subseteq \mathbb{R}$ be a bijection with inverse $g^{-1}: D' \to D$. Let the functional $T: \mathcal{F} \to \mathcal{P}(D), F \mapsto T(F) \subseteq D, D \subseteq \mathbb{R}$ be defined on a class $\mathcal{F}$ of probability distributions with CDF $F$. Let $\mathcal{F}^{(g)} \subseteq \mathcal{F}$ denote the subclass of the probability distributions in $\mathcal{F}$ which are such that $\underline{y}^{(g)} = g(\underline{y})$ has CDF $F^{(g)}$ and $F^{(g)} \in \mathcal{F}$. Define the functionals $T^{(g)}: \mathcal{F}^{(g)} \to \mathcal{P}(D), F^{(g)} \mapsto T^{(g)}(F^{(g)}) = T(F^{(g)})$, $T_g: \mathcal{F} \to \mathcal{P}(D), F \mapsto T_g(F) = g^{-1}(T(F))$ and $T_g^{(g)}: \mathcal{F}^{(g)} \to \mathcal{P}(D), F^{(g)} \mapsto T_g^{(g)}(F^{(g)}) = g^{-1}(T(F^{(g)}))$.

Then the following holds:

a. $T$ is elicitable, if and only if $T_g^{(g)}(F^{(g)})$ is elicitable.

b. $T$ is identifiable, if and only if $T_g^{(g)}(F^{(g)})$ is identifiable.

c. A function $S: D' \times D' \to \mathbb{R}, (z, y) \mapsto S(z, y)$, is $\mathcal{F}$-consistent for $T$, if and only if the function $S_g: D \times D \to \mathbb{R}, (z, y) \mapsto S_g(z, y) = S(g(z), g(y))$ is $\mathcal{F}^{(g)}$-consistent for $T_g^{(g)}(F^{(g)})$.

d. A function $S: D' \times D' \to \mathbb{R}, (z, y) \mapsto S(z, y)$, is strictly $\mathcal{F}$-consistent for $T$, if and only if the function $S_g: D \times D \to \mathbb{R}, (z, y) \mapsto S_g(z, y) = S(g(z), g(y))$ is strictly $\mathcal{F}^{(g)}$-consistent for $T_g^{(g)}(F^{(g)})$.

e. A function $V: D' \times D' \to \mathbb{R}, (z, y) \mapsto V(z, y)$ is an $\mathcal{F}$-identification function for $T$, if and only if the function $V_g: D \times D \to \mathbb{R}, (z, y) \mapsto V_g(z, y) = V(g(z), g(y))$ is an $\mathcal{F}^{(g)}$-identification function for $T_g^{(g)}(F^{(g)})$.

f. A function $V: D' \times D' \to \mathbb{R}, (z, y) \mapsto V(z, y)$ is a strict $\mathcal{F}$-identification function for $T$, if and only if the function $V_g: D \times D \to \mathbb{R}, (z, y) \mapsto V_g(z, y) = V(g(z), g(y))$ is a strict $\mathcal{F}^{(g)}$-identification function for $T_g^{(g)}(F^{(g)})$.■

The next example demonstrates an application of Theorem 4 for the case where the log transformation is applied to the prediction and realization variables of the loss function. It illustrates how a new strictly consistent loss function arises from the squared error loss function and how a new elicitable functional is constructed from the mean functional.

**Running example 2.2**: Let $g: \mathbb{R}^+ \to \mathbb{R}, t \mapsto g(t) = \log(t)$. The function $g$ is a bijection with inverse $g^{-1}: \mathbb{R} \to \mathbb{R}^+, t \mapsto g(t) = \exp(t)$. Let the functional $T: \mathcal{F} \to \mathbb{R}, F \mapsto T(F) = \mathbb{E}_F[\underline{y}] \subseteq \mathbb{R}$ be defined on the class $\mathcal{F}$ of probability distributions with finite second moment. Let $\mathcal{F}^{(g)} \subseteq \mathcal{F}$ denote the class of the probability distributions in $\mathcal{F}$ which are such that $\underline{y}^{(g)} = \log(\underline{y})$ has CDF $F^{(g)} \in \mathcal{F}^{(g)}$. $\mathcal{F}^{(g)}$ is the class of probability distributions $F^{(g)}$ of $\underline{y}^{(g)}$, with finite expectation $\mathbb{E}_{F^{(g)}}[\underline{y}^{(g)}]$ or equivalently finite expectation $\mathbb{E}_F[\log(\underline{y})]$. Define the functionals $T^{(g)}: \mathcal{F}^{(g)} \to \mathbb{R}, F \mapsto T^{(g)}(F^{(g)}) = \mathbb{E}_F[\log(\underline{y})] \subseteq \mathbb{R}$, $T_g: \mathcal{F} \to \mathcal{P}(D), F \mapsto T_g(F) = \exp(\mathbb{E}_F[\underline{y}])$ and $T_g^{(g)}: \mathcal{F}^{(g)} \to \mathbb{R}, F^{(g)} \mapsto T_g^{(g)}(F^{(g)}) = \exp(\mathbb{E}_F[\log(\underline{y})])$.

The loss function $S_{\text{SE}}: \mathbb{R} \times \mathbb{R} \to [0, \infty), (z, y) \mapsto S_{\text{SE}}(z, y) = (z - y)^2$ is strictly $\mathcal{F}$-consistent for the functional $T$; therefore, the loss function $S_g: \mathbb{R}^+ \times \mathbb{R}^+ \to [0, \infty), (z, y) \mapsto S_g(z, y) = (\log(z) - \log(y))^2$ is strictly $\mathcal{F}^{(g)}$-consistent for the functional $g^{-1}(T(F)) = \exp(\mathbb{E}_F[\log(\underline{y})])$.■

## 4.2 Applications to specific classes of consistent loss functions

### *4.2.1 Strictly consistent loss functions for $g$-transformed expectations*

The most widely used family of loss functions for transformed predictions and realizations assumes the form $S_g(z, y) = (g(z) - g(y))^2$, where $g$ is a bijection. $S_g$ represents a transformed version of the squared error loss function. By applying Remark 1, it follows that $S_g$ is a strictly consistent loss function that elicits the functional $g^{-1}(\mathbb{E}_F[g(\underline{y})])$, which we refer to as $g$-transformed expectation. Table 1 presents a range of choices for the function $g$, along with their respective elicitable functionals. Specific transformations such as $g(t) = \log(t), \log(t + b), t^a, a = -2, -1, -0.5, 0.2, 0.5, 2, (t + c)^{-1}$ and $(t^{0.25} - 1)/0.25$ (the Box-Cox transformation) have been empirically studied in the environmental science and technology literature ([8], [9]). The choice $g(t) = \exp(at)$, which is a special case of the Bregman function for $\varphi(t) = t^2$, was proposed by [41] and elicits the entropic risk measure [42] $\log(\mathbb{E}_F[\exp(a\underline{y})])/a$, when employed in the squared error loss function.

By applying Remark 1, it follows that the function $V_g(z, y) = g(z) - g(y)$ is a strict identification function for the functional $g^{-1}(\mathbb{E}_F[g(\underline{y})])$. Identifiable functionals of this form, along with their corresponding strict identification functions, are also presented in Table 1, for a range of choices of $g$.

Table 1. Elicitable and identifiable functionals corresponding to the strictly consistent loss function $S_g(z,y) = (g(z) - g(y))^2$ and the strict identification function $V_g(z,y) = g(z) - g(y)$, respectively.

| Function $g(t)$ | Function $g^{-1}(t)$ | Elicitable – identifiable functional |
|---|---|---|
| $g: \mathbb{R}^+ \to \mathbb{R}, t \mapsto g(t) = \log(t)$ | $g^{-1}(t) = \exp(t)$ | $\exp(\mathbb{E}_F[\log(\underline{y})])$ |
| $g: (-b/a, \infty) \to \mathbb{R}, t \mapsto g(t) = \log(at+b), a > 0$ | $g^{-1}(t) = (\exp(t) - b)/a$ | $(\exp(\mathbb{E}_F[\log(a\underline{y}+b)]) - b)/a$ |
| $g: (-b, \infty) \to \mathbb{R}, t \mapsto g(t) = \log(t+b)$ | $g^{-1}(t) = \exp(t) - b$ | $\exp(\mathbb{E}_F[\log(\underline{y}+b)]) - b$ |
| $g: \mathbb{R} \to \mathbb{R}^+, t \mapsto g(t) = \exp(t)$ | $g^{-1}(t) = \log(t)$ | $\log(\mathbb{E}_F[\exp(\underline{y})])$ |
| $g: \mathbb{R} \to \mathbb{R}^+, t \mapsto g(t) = \exp(at+b), a \neq 0$ | $g^{-1}(t) = (\log(t) - b)/a$ | $(\log(\mathbb{E}_F[\exp(a\underline{y}+b)]) - b)/a$ |
| $g: [0, \infty) \to [0, \infty), t \mapsto g(t) = t^a, a > 0$ | $g^{-1}(t) = t^{1/a}$ | $(\mathbb{E}_F[\underline{y}^a])^{1/a}$ |
| $g: [-c/b, \infty) \to [0, \infty), t \mapsto g(t) = (bt+c)^a, a, b > 0$ | $g^{-1}(t) = (t^{1/a} - c)/b$ | $((\mathbb{E}_F[(b\underline{y}+c)^a])^{1/a} - c)/b$ |
| $g: \mathbb{R}^+ \to \mathbb{R}^+, t \mapsto g(t) = t^a, a \neq 0$ | $g^{-1}(t) = t^{1/a}$ | $(\mathbb{E}_F[\underline{y}^a])^{1/a}$ |
| $g: (-c/b, \infty) \to \mathbb{R}^+, t \mapsto g(t) = (bt+c)^a, a \neq 0, b > 0$ | $g^{-1}(t) = (t^{1/a} - c)/b$ | $((\mathbb{E}_F[(b\underline{y}+c)^a])^{1/a} - c)/b$ |
| $g: (-c, \infty) \to \mathbb{R}^+, t \mapsto g(t) = (t+c)^a, a \neq 0$ | $g^{-1}(t) = t^{1/a} - c$ | $(\mathbb{E}_F[(\underline{y}+c)^a])^{1/a} - c$ |
| $g: \mathbb{R}^+ \to \begin{cases} (-1/a, \infty), a > 0 \\ \mathbb{R}, a = 0 \\ (-\infty, -1/a), a < 0 \end{cases}, t \mapsto g(t) = \begin{cases} (t^a - 1)/a, a \neq 0 \\ \log(t), a = 0 \end{cases}$ | $g^{-1}(t) = \begin{cases} (at+1)^{1/a} \\ \exp(t) \end{cases}$ | $\begin{cases} (\mathbb{E}_F[\underline{y}^a])^{1/a} \\ \exp(\mathbb{E}_F[\log(\underline{y})]) \end{cases}$ |

If $g$ is a strictly increasing, once differentiable function, then

$$\frac{\partial S_g(z,y)}{\partial z} = 2g'(z)V_g(z,y) \quad (4.1)$$

where $g'(z)$ is positive and $V_g$ is an oriented strict identification function. $V_g$ is a strict identification function because $\mathbb{E}_F[V_g(z,\underline{y})] = 0 \Leftrightarrow g(z) = \mathbb{E}_F[g(\underline{y})] \Leftrightarrow z = g^{-1}(\mathbb{E}_F[g(\underline{y})])$. It is an oriented function because $z > g^{-1}(\mathbb{E}_F[g(\underline{y})]) \Leftrightarrow g(z) > \mathbb{E}_F[g(\underline{y})] \Leftrightarrow \mathbb{E}_F[V_g(z,\underline{y})] > 0$. By applying Osband's principle (see Section 3.4), it follows that $S_g(z,y)$ is a strictly consistent loss function. If $g$ is a strictly decreasing, then $-g$ is strictly increasing, and $-V_g$ becomes oriented, enabling the same construction of the strictly consistent loss function $S_g$. This provides an alternative proof of parts of Remark 1, provided $S$ is restricted to squared error loss functions and $g$ to monotonic, once differentiable functions.

Functions of the form $(g(z) - g(y))^2$ constitute a subset of a broader class of loss functions that are strictly consistent for the functional $g^{-1}(\mathbb{E}_F[g(\underline{y})])$. Applying Remark 1 to the Bregman family of loss functions (3.44), it follows that a loss function is strictly consistent for the functional $g^{-1}(\mathbb{E}_F[g(\underline{y})])$ if and only if is of the form $S(z,y;\varphi,g) = \varphi(g(y)) - \varphi(g(z)) + \varphi'(g(z))(g(z) - g(y))$, where $\varphi$ is strictly convex function. The special case $S(z,y;\varphi,\exp(at))$ was introduced by [41].

#### *4.2.2 Explaining the functional $g^{-1}(\mathbb{E}_F[g(\underline{y})])$*

To clarify the distinction between the functionals $\mathbb{E}_F[\underline{y}]$ and $g^{-1}(\mathbb{E}_F[g(\underline{y})])$ we confine our analysis to the log-normal family of probability distributions for $\underline{y}$. Let $\underline{y} \sim \text{Lognormal}(\mu, \sigma^2)$, where $\mu \in \mathbb{R}$ and $\sigma > 0$, with support $(0, \infty)$. We then analyze the family of loss functions $S_g(z, y) = (z^a - y^a)^2$, which is strictly consistent for the functional $T(a) = \exp(\mu + a\sigma^2/2)$.

Notably, $T(a)$ is a strictly increasing function of $a$, which partially explains the findings of [9]. [9] explored loss functions of the form $(z^a - y^a)^2$, by varying the parameter $a$. They trained a hydrological (physics-based) model $\text{HM}(a)$ to predict river streamflow data using these loss functions with different values of $a$. The model's predictions were then evaluated across various streamflow ranges using the squared error loss function. [9] concluded that as $a$ increases, the model's predictions improve for higher streamflows. This can be explained, under the assumption of a log-normal distribution for streamflow data [43], and by the fact that the functional $T(a)$, which the models aims to predict, increases with $a$.

*4.2.3 The median functional*

Our previous analysis has focused on $g$-transformed expectations. In such cases, Remark 1 leverages LOTUS to map the functional from the transformed distribution $F^{(g)}$ back to the original distribution $F$. For the median functional, however, Remark 1 applies a distinct mechanism. Specifically, if $g$ is a monotonic function, the median satisfies the property $Q^{1/2}\big(F^{(g)}\big) = g(Q^{1/2}(F))$ implying that $g^{-1}(Q^{1/2}(F^{(g)})) = Q^{1/2}(F)$. Here $g^{-1}$ directly links the median of the transformed distribution $F^{(g)}$ to the median of $F$, contrasting with $g$-transformed expectations.

By applying the transformation $g$ to the variables of the absolute error loss function (3.14) we obtain $S_{\text{AE}}(g(z), g(y)) = |g(z) - g(y)|$, which, by Remark 1, is strictly consistent for the median functional. This loss function $S_{\text{AE}}(g(z), g(y))$ corresponds to a special case of the GPL loss function (3.32) for $\tau = 1/2$ up to a multiplicative constant. Additionally, Remark 1 shows that the function $V_{Q^{1/2}}(g(z), g(y)) = \mathbb{1}(g(z) \geq g(y)) - 1/2$, constructed from the strict quantile identification function in eq. (3.33) for $\tau = 1/2$, acts as a strict identification function for the median functional. It is straightforward to show that $V_{Q^{1/2}}(g(z), g(y)) \propto V_{Q^{1/2}}(z, y)$, regardless of whether $g$ is strictly increasing or strictly decreasing.

*4.2.4 Quantile functionals*

The quantile functional presents a more complex case, as the monotonicity type of $g$ determines how quantiles of the transformed distribution $F^{(g)}$ map back to the original distribution $F$:

a. If $g$ is strictly increasing, $Q^{\tau}\big(F^{(g)}\big) = g(Q^{\tau}(F))$, implying $g^{-1}(Q^{\tau}(F^{(g)})) = Q^{\tau}(F)$.

b. If $g$ is strictly decreasing, $Q^{\tau}\big(F^{(g)}\big) = g(Q^{1-\tau}(F))$, implying $g^{-1}(Q^{\tau}(F^{(g)})) = Q^{1-\tau}(F)$.

For strictly increasing $g$, applying the transformation to the variables of the asymmetric piecewise linear loss function (3.34), yields the GPL loss function (3.32), since $\mathbb{1}\big(g(z) \geq g(y)\big) = \mathbb{1}(z \geq \mathrm{y})$. Similarly, transforming the quantile strict identification function (3.33) via $g$ results in $V_{Q^{\tau}}(g(z), g(y)) = V_{Q^{\tau}}(z, y)$.

For strictly decreasing $g$, the relationship is less straightforward. Notably, $\mathbb{1}\big(g(z) \geq g(y)\big) - \tau = -(\mathbb{1}(z \geq \mathrm{y}) - (1 - \tau))$. Applying $g$ to the asymmetric piecewise linear loss function (3.34) produces $S_Q(g(z), g(y); \tau, t) = (\mathbb{1}\big(g(z) \geq g(y)\big) - \tau)\big(g(z) - g(y)\big)$ or equivalently, $S_Q(g(z), g(y); \tau, t) = (\mathbb{1}(z \geq \mathrm{y}) - (1 - \tau))(-g(z) - (-g(y)))$. This loss function is strictly consistent for the functional $Q^{1-\tau}(F)$, representing a special case of the GPL loss function, where $-g$ is strictly increasing.

Regardless of whether $g$ is increasing or decreasing, Remark 1 demonstrates how the asymmetric piecewise-linear loss function generalizes to the broader class of GPL loss functions. While this connection is established (see Section 3.8.1 and [44]), this section arrives at this relationship through a different analytical path, specifically by analyzing how quantile loss functions transform.

*4.2.5 Strictly consistent loss functions for expectile-based functionals*

Analogous to $g$-transformed expectations, elicitable functionals can also be constructed using expectiles, which we refer to as $g$-transformed expectiles. Applying Remark 1 to the expectile loss function (3.40) results in the strictly consistent loss function:

$$S_E(z, y; \tau, \varphi, g) = |\mathbb{1}(g(z) \geq g(y)) - \tau|(\varphi(g(y)) - \varphi(g(z)) + \varphi'(g(z))(g(z) - g(y))) \quad (4.2)$$

To identify the respective elicitable functional, we observe that $E^{\tau}\big(F^{(g)}\big) = \{z \in I: \tau \mathbb{E}_{F^{(g)}}[\kappa_{0,\infty}(\underline{y} - z)] = (1 - \tau)\mathbb{E}_{F^{(g)}}[\kappa_{0,\infty}(z - \underline{y})\}$. By LOTUS, this becomes $E^{\tau}\big(F^{(g)}\big) =$ $\{z \in I: \tau \mathbb{E}_F[\kappa_{0,\infty}(g(\underline{y}) - z)] = (1-\tau)\mathbb{E}_F[\kappa_{0,\infty}(z - g(\underline{y}))\}$. Consequently, the loss function (4.2) elicits the $g$-transformed expectile:

$$g^{-1}(E^\tau(F^{(g)})) = g^{-1}(\{z \in I: \tau \mathbb{E}_F[\kappa_{0,\infty}(g(\underline{y}) - z)] = (1-\tau)\mathbb{E}_F[\kappa_{0,\infty}(z - g(\underline{y}))\}) \quad (4.3)$$

The $g$-transformed expectation arises as a special case of the $g$-transformed expectile, when $\tau = 1/2$. While this follows directly from the fact that strictly consistent loss functions for $g$-transformed expectations are special cases of the strictly loss function (4.2) with $\tau = 1/2$, we present an alternative proof for clarity. Define $A = \{z \in I: \mathbb{E}_F[\kappa_{0,\infty}(g(\underline{y}) - z) - \kappa_{0,\infty}(z - g(\underline{y}))] = 0\}$. Rewriting $A$ we have $A = \{z \in I: \mathbb{E}_F[(g(\underline{y}) - z)\, \mathbb{1}(g(\underline{y}) \geq z) - (z - g(\underline{y}))\, \mathbb{1}(g(\underline{y}) \leq z)] = 0\}$, which simplifies to $A = \{z \in I: \mathbb{E}_F[g(\underline{y})\, \mathbb{1}(g(\underline{y}) \geq z) + g(\underline{y})\, \mathbb{1}(g(\underline{y}) \leq z)] = \mathbb{E}_F[z\, \mathbb{1}(g(\underline{y}) \geq z) + z\, \mathbb{1}(g(\underline{y}) \leq z)]\}$. The left-hand expectation reduces to $\mathbb{E}_F[g(\underline{y})]$, while the right-hand expectation equals $z$. Hence, $A = \mathbb{E}_F[g(\underline{y})]$, and it follows that $g^{-1}(E^{1/2}(F^{(g)})) = g^{-1}(\mathbb{E}_F[g(\underline{y})])$ completing the proof.

The $g$-transformed expectile (4.3) is also identifiable. Its strict identification function constructed by applying Remark 1 to the expectile strict identification function (3.41) is $V_{E^\tau}(g(z), g(y)) = 2|\mathbb{1}(g(z) \geq g(y)) - \tau|(g(z) - g(y))$.

*4.2.6 The pair (mean, variance) of transformed realizations*

Theorems 3 and 4 generalize naturally to $k$-dimensional functionals. As in illustration, consider the pair (mean, variance). The strict identification function (3.45) and strictly consistent loss function (3.46) assume the following forms respectively:

$$V_{\mathrm{mv}}(x_1, x_2, g(y)) = (x_1 - g(y), x_2 + x_1^2 - g^2(y)) \quad (4.4)$$

$$S_{\mathrm{mv}}(x_1, x_2, g(y)) = x_2^{-2}(x_1^2 - 2x_2 - 2x_1 g(y) + g^2(y)) \quad (4.5)$$

By theorems 3 and 4, the functions (4.4) and (4.5) respectively identify and elicit the pair $(\mathbb{E}_F[g(\underline{y})], \mathrm{Var}_F[g(\underline{y})])$.

*4.2.7 Skill scores*

Skill scores of the form (3.25) employing a climatology reference method, can be directly constructed for loss functions involving transformed variables. Consider, for example the loss functions discussed in Section 4.2.1. The climatology for the functional is defined via the inverse transformation $g^{-1}((1/n)\sum_{i=1}^{n} g(y_i))$, leading to predictions of the reference method given by $\mathbf{z}_{\text{ref}} = (1/n)\mathbf{1}g^{-1}((1/n)\sum_{i=1}^{n} g(y_i))$.

## 5. Applications

To demonstrate the implications of variable transformations in consistent loss functions, we examine three illustrative examples. The first case (Section 5.1) explores these implications through parameter estimation for a probability distribution, analyzed via a simulation study. Section 5.2 extends the simulation framework from Section 5.1 by investigating the effects of these transformations in a regression context. Finally, Section 5.3 evaluates a real-world regression application to further substantiate the practical significance of the findings. Supplementary materials include reproducible code and datasets for all analyses in this section. All computations were performed using the `R` programming language ([45], version 4.5.0) within the `RStudio` integrated development environment (version 2024.12.1+563).The contributed packages used for these analyses are listed in Appendix C.

### 5.1 Estimation of probability distribution parameters

This example serves to computationally demonstrate the consistency of the loss function $S(z, y) = (z^a - y^a)^2$ (proven theoretically in Section 4.2.1) within a controlled $M$-estimation framework (Section 3.7). As discussed in Section 3.7, consistency equivalence holds between parameter estimation and point prediction evaluation [12].

To demonstrate the impact of loss function selection on parameter estimation for probability distributions, we simulate data from random variable $\underline{y} \sim \text{Lognormal}(\mu, \sigma)$ that follows the log-normal distribution with probability density function (PDF)

$$f(y; \mu, \sigma) := \frac{1}{y\,\sigma\sqrt{2\pi}} \exp\left(-\frac{(\log y - \mu)^2}{2\sigma^2}\right), y > 0, \mu \in \mathbb{R}, \sigma > 0 \tag{5.1}$$

A sample of $n = 1\,000$ realizations was simulated with true parameters $\mu = 0$ and $\sigma = 1.2$. Figure 2 shows the histogram of the simulated data. Maximum likelihood estimation yielded parameter estimates $\hat{\mu} = 0.055$ and $\hat{\sigma} = 1.198$.

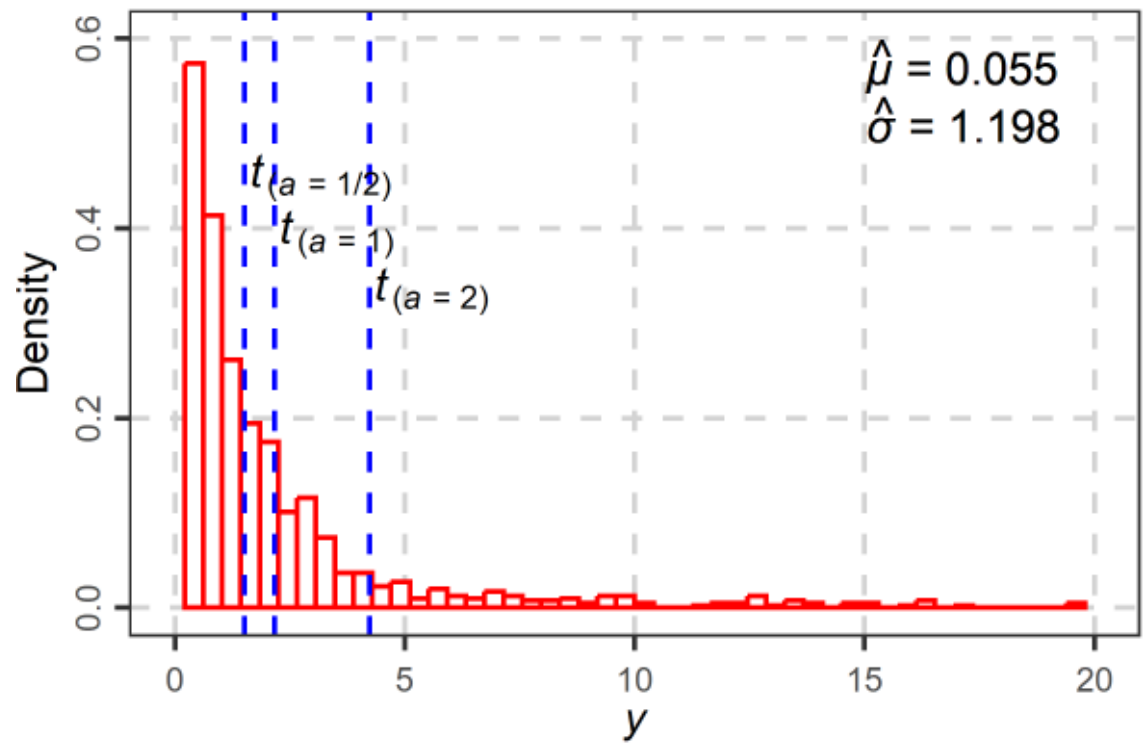

Figure 2. Histogram of 1 000 simulated values from a log-normal distribution ($\mu = 0, \sigma = 1.2$). Maximum likelihood estimates from the sample are $\hat{\mu} = 0.055$ and $\hat{\sigma} = 1.198$. Blue-dashed lines indicate estimates $\hat{t}(a)$ of the parameter $t(a) = \exp(\mu + a\sigma^2/2)$, for $a = 1/2, 1, 2$, computed by minimizing the average score $\bar{S}(t, \boldsymbol{y}) = (1/n)\sum_{i=1}^{n}(t^a - y_i^a)^2$.

The support of the log-normal distribution is the set of positive numbers. Therefore, by minimizing $\bar{S}(t, \boldsymbol{y}) = (1/n)\sum_{i=1}^{n} S(t, y_i)$, where $S(z, y) = (z^a - y^a)^2$, we estimate the parameter $t(a) = (\mathbb{E}_F[\underline{y}^a])^{1/a}$ (see Table 1). For the log-normal distribution, this parameter simplifies to $t(a) = \exp(\mu + a\sigma^2/2)$. Estimates $\hat{t}(a)$ of $t(a)$, for $a = 1/2, 1, 2$ are shown in Figure 2. The true values of $t(a)$ are $1.43, 2.05, 4.22$, while the estimates obtained by minimizing $\bar{S}$ are $\hat{t}(a) = 1.51, 2.16, 4.23$ respectively.

## 5.2 Regression with simulated data

The previous example illustrated the role of consistent loss functions in parameter estimation. We now extend this framework to semiparametric models. For this purpose, we simulated $n = 2\,000$ realizations of:

(a) A predictor $\underline{x} \sim \text{Lognormal}(\mu_{\underline{x}}, \sigma_{\underline{x}})$ with $\mu_{\underline{x}} = 0$ and $\sigma_{\underline{x}} = 1$,

(b) An error term $\underline{\varepsilon} \sim \text{Lognormal}(\mu_{\underline{\varepsilon}}, \sigma_{\underline{\varepsilon}})$ with $\mu_{\underline{\varepsilon}} = 0$ and $\sigma_{\underline{\varepsilon}} = 1$.

Subsequently, we simulated 2 000 realizations of the dependent variable $\underline{y} = d_0 \underline{x}^{c_0} \underline{\varepsilon}$ with $c_0 = 0.9$ and $b_0 = 0.1$. The variable $\underline{y}$ follows a log-normal distribution with parameters $\mu_{\underline{y}} = \log(d_0) + c_0\mu_{\underline{x}} + \mu_{\underline{\varepsilon}}$ and $\sigma_{\underline{y}} = (c_0^2\sigma_{\underline{x}}^2 + \sigma_{\underline{\varepsilon}}^2)^{1/2}$ (see Figure 3a). Conditional on a realization $x$, the random variable $\underline{y}|x$ is also log-normally distributed with parameters $\mu_{\underline{y}|x} = \log(d_0) + c_0 \log(x) + \mu_{\underline{\varepsilon}}$ and $\sigma_{\underline{y}|x} = \sigma_{\underline{\varepsilon}}$.

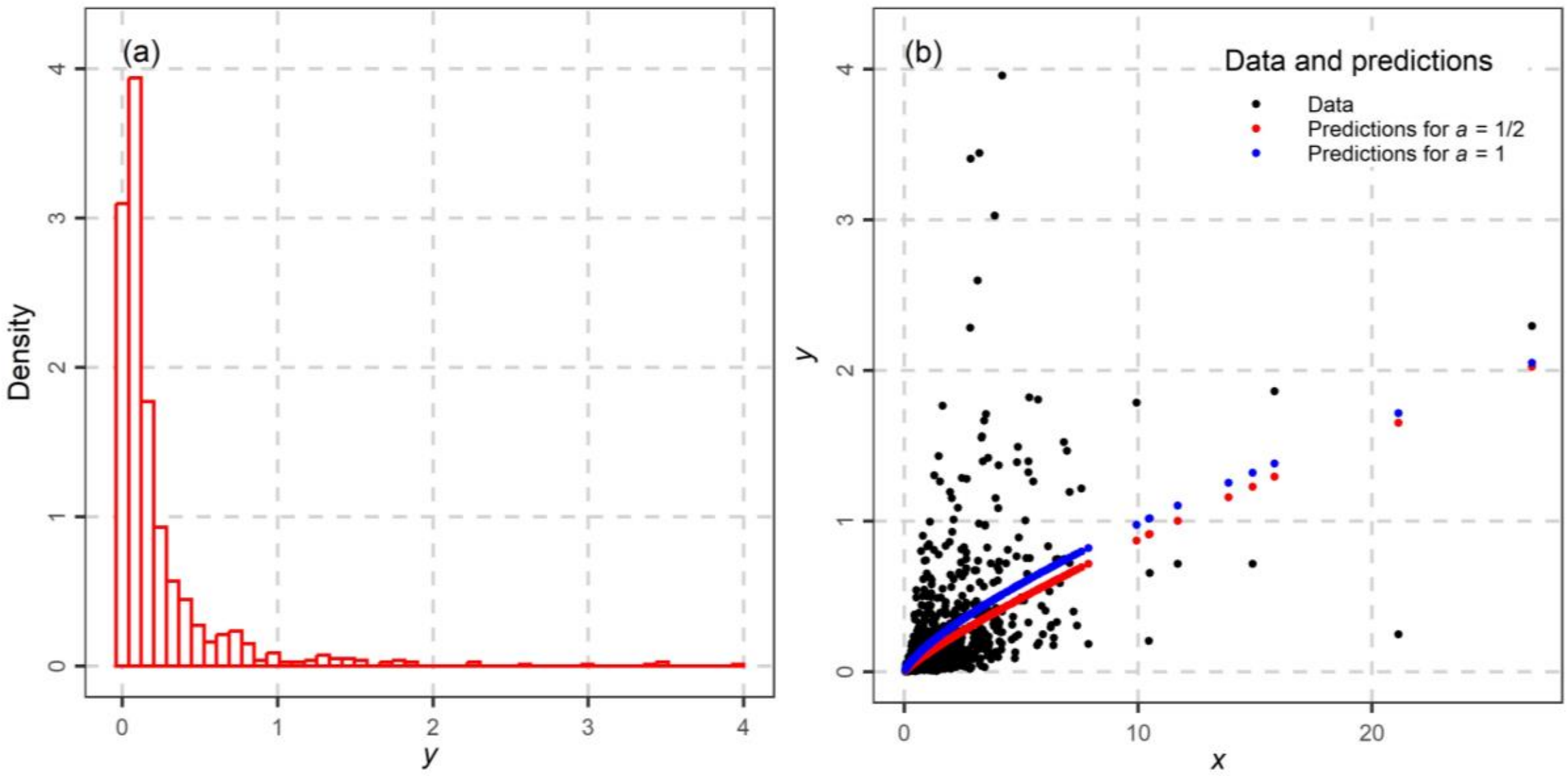

Figure 3. (a) Histogram of the dependent variable $\underline{y} = d_0 \underline{x}^{c_0} \underline{\varepsilon}$ for the 1 000 test set realizations. (b) Test set data (predictor $x$ vs. dependent variable $y$) with predictions from the semiparametric model $M(a)$ model, trained using the loss function $S(z, y) = (z^a - y^a)^2$ for $a = 1/2$ and $a = 1$.

We define the semiparametric model $M(a)$: $z = dx^c$, where $\boldsymbol{\theta} = (c, d)$ are the model parameters. The dataset is split into training and test sets, each containing 1 000 realizations of $\underline{x}$ and $\underline{y}$. The goal of the semi-parametric model is to predict the functional $T(F_{\underline{y}|\underline{x}})$ conditional on the predictor $x$ as outlined in Section 3.7). To estimate $\boldsymbol{\theta}$, we trained $M(a)$ on the training set using the loss function $S(z, y) = (z^a - y^a)^2$ for $a = 1/2, 1$. As shown in Section 3.7, the estimator $\widehat{\boldsymbol{\theta}}_l = \arg\min_{\boldsymbol{\theta} \in \boldsymbol{\Theta}} (1/l) \sum_{i=1}^{l} S(dx_i^c, y_i)$ is consistent for $\boldsymbol{\theta}$.

Figure 3b displays the test set data and model predictions for $a = 1/2$ and $a = 1$ in the test set. Since $\underline{y}|x$ is log-normally distributed (thus with support the set of positive numbers), Table 1 confirms that $S$ is consistent for the functional $T(F_{\underline{y}|\underline{x}}) = \exp(\mu + a\sigma^2/2)$. Thus training $M(a)$ with $S(z, y) = (z^a - y^a)^2$, produces predictions targeting $\exp(\mu + a\sigma^2/2)$. Predictions for $a = 1$ are systemically higher than those for $a = 1/2$, consistent with the fact that $\exp(\mu + a\sigma^2/2)$ increases with $a$.

## 5.3 A regression application to real-world data

To complement our simulation experiments, we analyze real-world data from the Melbourne housing market following the application by [46]. The dataset, sourced from Kaggle (https://www.kaggle.com/datasets/anthonypino/melbourne-housing-market, date of access: 2023-04-02) originally contains 34 857 property records. After excluding entries with incomplete predictor data or zero building area, we retained a refined subset of 8 798 houses.

In this regression framework, building area serves as the predictor variable for house price. Both variables are positive, and their histograms (Figures 4a and 4b) exhibit right-skewed distributions consistent with a log-normal probability (a finding aligned with [46] for house prices). Following the methodology in Section 5.2, we train the semiparametric model $M(a)$: $z = dx^c$, where $x$ represents building area, $z$ is the predicted functional of the conditional distribution $F_{\underline{y}|\underline{x}}$ and $\boldsymbol{\theta} = (c, d)$ are model parameters. As outlined in Section 3, we distinguish between actual house price values $y$, and the model's predicted functional $z = T(F_{\underline{y}|\underline{x}})$, which targets a property of the conditional distribution rather than direct prediction of $y$.

The dataset is split equally into training and test sets ($n = 4\,399$ each). We estimate $\boldsymbol{\theta}$ by minimizing the average loss $(1/n)\sum_{i=1}^{n} S(dx_i^c, y_i)$, where $S(z, y) = (z^a - y^a)^2$ for $a = 1/2, 1, 2$. Figure 4c displays test set data and model predictions for $a = 1/2$, $a = 1$ and $a = 2$. Predictions systematically increase with $a$, mirroring the simulation results in Section 5.2. These results show that predictions for $a = 2$ exceed those for $a = 1$, which in turn exceed predictions for $a = 1/2$.

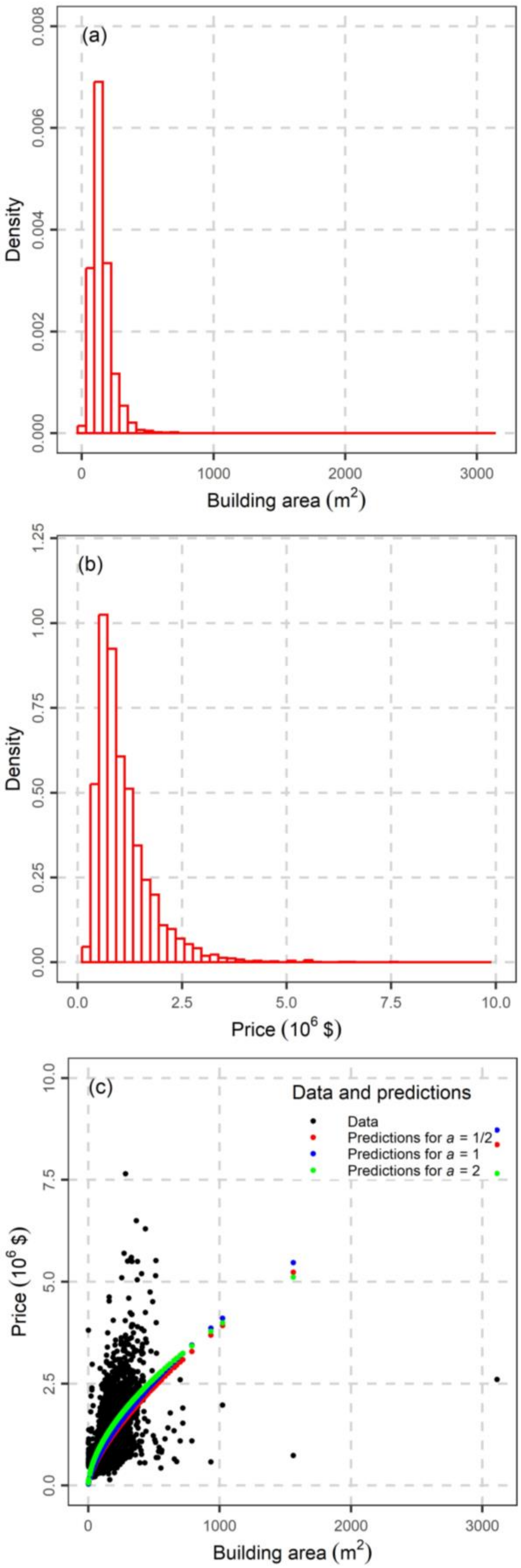

Figure 4. (a) Histogram of building area values in Melbourne. (b) Histogram of house prices. (c) Test set data with predictions from the semiparametric model $M(a)$ model, trained using the loss function $S(z, y) = (z^a - y^a)^2$ for $a = 1/2$, $a = 1$ and $a = 2$.

For each trained model $M(a)$ we compute the average test set score for all loss function types (Table 2). As shown in Table 2, a model trained with a pre-specified loss function, achieves superior performance on that specific loss metric compared to models trained with other loss functions. For example, model $M(a = 1/2)$ outperforms other models when evaluated on $S(z, y) = (z^{1/2} - y^{1/2})^2$, the loss function it was optimized for. This aligns with the theoretical framework of [16], as discussed in Section 1, and underscores two critical points:

(a) The need to pre-specify both the target functional and the loss function used for training and evaluation.

(b) The methodological inconsistency of evaluating model performance with multiple competing metrics, which is a common practice that lacks theoretical grounding.

Table 2. Average test set scores for each trained model $M(a)$ and loss function $S(z, y) = (z^a - y^a)^2$ $(a = 1/2, 1, 2)$. Best performances for each loss type are highlighted in bold.

| | Average score | | |
|---|---|---|---|
| Model | $\sum_{i=1}^{n} (z_i^{1/2} - y_i^{1/2})^2 /n$ | $\sum_{i=1}^{n} (z_i - y_i)^2/n$ | $\sum_{i=1}^{n} (z_i^2 - y_i^2)^2/n$ |
| $M(a = 1/2)$ | **0.0556** | 0.3361 | 7.7513 |
| $M(a = 1)$ | 0.0562 | **0.3334** | 7.8139 |
| $M(a = 2)$ | 0.0659 | 0.3555 | **7.0824** |

## 6. Discussion and future outlook

This study is motivated by research in environmental science and technology, where loss functions of the form $S(g(z), g(y))$ are widely used for training predictive models. Although the extant literature emphasizes that predictions should be probabilistic ([47], [48]), most practical applications still employ loss functions to generate point predictions. In hydrology (a major subdiscipline of environmental science), functionals of hydrologic processes are termed hydrograph functionals [48]. Building on this framework, we developed novel functionals using such loss functions and characterized their elicitability and identifiability. Our findings advance the understanding of model training methodologies, providing theoretical insights into optimizing scoring mechanisms for environmental science applications. These contributions complement existing environmental methods that train models using squared-error [29], quantile ([49], [50], [51]) and expectile [52] loss functions.

Environmental predictions from models trained with functions of the form $(g(z) - g(y))^2$ were explained under the assumption of a log-normal distribution for the environmental process. However, such loss functions are strictly consistent estimators for broader classes of probability distributions. Consequently, the empirical interpretation of model predictions can be extended to include distributions that better represent environmental processes.

While our work focuses on transformations of quantile, expectile, and mean-variance strictly consistent loss functions, the results are general in scope. They can thus be applied to broader classes of consistent loss and identification functions, including those involving multi-dimensional variables and functionals. Additionally, new proper scoring rules can be constructed from the proposed consistent loss functions [2].

While new functionals such as the $g$-transformed expectation and $g$-transformed expectile are introduced here, broader adoption of these tools would benefit from a clearer economic interpretation. Existing research on quantiles and expectiles [53] provides a foundation, but future work could explore how their framework applies to these novel functionals. Given that such functionals can serve as risk measures, understanding their properties (such as monotonicity, coherence, and other axiomatic characteristics) is critical. For example, debates persist about the utility of expectiles compared to quantiles in risk modeling [54]. Extending statistical properties, such as symmetry testing in nonparametric regression (analogous to the case of $L_p$-quantiles studied by [55]), could provide another path forward for analyzing these new functionals.

## 7. Conclusions

In this study, we analyzed loss functions constructed through transformations of the realization variable in strictly consistent loss functions. We proved that these transformed functions retain strict consistency for a corresponding transformation of the elicitable functional linked to the original loss function. By extending Osband's revelation principle, we characterized cases where both the realization and prediction variables of the original loss functions are transformed via a bijection $g$. Similar logic applies to transformations of strict identification functions, enabling analogous constructions. These characterizations establish a systematic approach for generating new identifiable and/or elicitable functionals.

When applied to Bregman and expectile loss functions, this methodology produced strictly consistent loss functions for two novel functionals: the $g$-transformed expectation and $g$-transformed expectile. Consistent loss functions for $g$-transformed expectations have already been empirically validated in environmental science and technology applications. Here, we interpreted these empirical findings through the theoretical properties of the proposed functionals. Additionally, our analysis of quantile loss functions uncovered relationships between the asymmetric piecewise linear loss function and the generalized piecewise linear (GPL) loss function.

The generality of this framework opens new ways for designing elicitable and identifiable functionals with wide-ranging applicability. By bridging theoretical insights with practical tools, this work advances the adaptability of loss function transformations in machine learning methodology and interdisciplinary research.

## Appendix A Vector notation

The notation of vectors remains consistent throughout the manuscript. We write a vector with $n$ elements as

$$\boldsymbol{x} = (x_1, \dots, x_n)^{\mathrm{T}} \tag{A.1}$$

The zero and one vectors are denoted by:

$$\mathbf{0} = (0, \dots, 0)^{\mathrm{T}} \tag{A.2}$$

$$\mathbf{1} = (1, \dots, 1)^{\mathrm{T}} \tag{A.3}$$

The sample mean (mean climatology) of a vector is defined by [56, p. 45]

$$\overline{\boldsymbol{x}} := (1/n)\mathbf{1}^{\mathrm{T}}\boldsymbol{x} = (1/n)\sum_{i=1}^{n} x_i = (1/n)\langle \boldsymbol{x}, \mathbf{1} \rangle \tag{A.4}$$

where $\langle \cdot,\cdot \rangle$ denotes the inner product of two vectors.

## Appendix B Proofs

**Proof of Theorem 3**: We first prove part (b). Let $F \in \mathcal{F}^{(g)}$, $t^{(g)} \in T^{(g)}(F)$, and $\boldsymbol{z} \in \mathbb{R}^k$. Then $\mathbb{E}_F[S^{(g)}(t^{(g)}, \underline{\boldsymbol{y}})] = \mathbb{E}_F[S(t^{(g)}, g(\underline{\boldsymbol{y}}))] = \mathbb{E}_{F^{(g)}}[S(t^{(g)}, \underline{\boldsymbol{y}}^{(g)})] \leq \mathbb{E}_{F^{(g)}}[S(\boldsymbol{z}, \underline{\boldsymbol{y}}^{(g)})] = \mathbb{E}_F[S(\boldsymbol{z}, g(\underline{\boldsymbol{y}}))] = \mathbb{E}_F[S^{(g)}(\boldsymbol{z}, \underline{\boldsymbol{y}})]$. The critical inequality holds because $\underline{\boldsymbol{y}}^{(g)}$ has CDF $F^{(g)}$, $F^{(g)} \in \mathcal{F}^{(g)} \subseteq \mathcal{F}$ and $t^{(g)} \in T^{(g)}(F) = T(F^{(g)})$. The equality $\mathbb{E}_F[S(t^{(g)}, g(\underline{\boldsymbol{y}}))] = \mathbb{E}_{F^{(g)}}[S(t^{(g)}, \underline{\boldsymbol{y}}^{(g)})]$ is a consequence of the law of the unconscious statistician.

To prove parts (c) and (a), we note that the inequality is strict if $S$ is strictly $\mathcal{F}$-consistent for $T$, unless $\boldsymbol{z} \in T^{(g)}(F) = T(F^{(g)})$.

Part (d) is established by observing that $g$ is a bijection and by applying the implications demonstrated in parts (a), (b) and (c).■

**Proof of Theorem 4**: We first prove part (b). Let $F^{(g)} \in \mathcal{F}^{(g)}$, $t^{(g)} \in T^{(g)}(F)$, and $\boldsymbol{z} \in \mathbb{R}^k$. Then $\mathbb{E}_F[V^{(g)}(t^{(g)}, \underline{\boldsymbol{y}})] = \mathbb{E}_F[V(t^{(g)}, g(\underline{\boldsymbol{y}}))] = \mathbb{E}_{F^{(g)}}[V(t^{(g)}, \underline{\boldsymbol{y}}^{(g)})] = 0$. The critical equality to zero holds because $\underline{\boldsymbol{y}}^{(g)}$ has CDF $F^{(g)}$, $F^{(g)} \in \mathcal{F}^{(g)} \subseteq \mathcal{F}$ and $t^{(g)} \in T^{(g)}(F) = T(F^{(g)})$. The equality $\mathbb{E}_F[V(t^{(g)}, g(\underline{\boldsymbol{y}}))] = \mathbb{E}_{F^{(g)}}[V(t^{(g)}, \underline{\boldsymbol{y}}^{(g)})]$ is a consequence of the law of the unconscious statistician.

To prove parts (c) and (a), we note that the equality to zero is strict if $S$ is a strict $\mathcal{F}$-identification function for $T$, unless $\boldsymbol{z} \in T^{(g)}(F) = T(F^{(g)})$.

Part (d) is established by observing that $g$ is a bijection and by applying the implications demonstrated in parts (a), (b) and (c).■

### Appendix C Statistical software

We used the `R` programming language [45] to implement the algorithms, and to report and visualize the results.

For data processing and visualizations, we used the contributed `R` packages `data.table` [57], `MASS` ([58], [59]), `tidyverse` ([60], [61]).

The performance metrics were computed by implementing the contributed `R` package `scoringfunctions` ([4], [62]).

Reports were produced by using the contributed `R` packages `devtools` [63], `knitr` ([64], [65], [66]), `rmarkdown` ([67], [68], [69]).

**Supplementary material:** We provide the following data and files to reproduce result of Section 5:

Section 5.1: `Simulation_example_1a.Rmd`, `Simulation_example_1a.html`, `Simulation_example_1a.RData`, `Simulation_example_1b_figures.Rmd`, `Simulation_example_1b_figures.html`, `Simulation_example_1b_figures.docx`.

Section 5.2: `Simulation_example_2a.Rmd`, `Simulation_example_2a.html`, `Simulation_example_2a.RData`, `Simulation_example_2b_figures.Rmd`, `Simulation_example_2b_figures.html`, `Simulation_example_2b_figures.docx`.

Section 5.3: Data folder `melbourne_housing`, `Melbourne_housing_final_data.RData`,

`real_example_1a_melbourne.Rmd`, `real_example_1a_melbourne.html`, `real_example_1a_melbourne.RData`, `real_example_1b_melbourne_figures.Rmd`, `real_example_1b_melbourne_figures.html`, `real_example_1b_melbourne_figures.docx`.

**Acknowledgement:** We thank the Associate Editor and reviewers whose comments contributed to improving the manuscript.

**Conflicts of interest:** The authors declare no conflict of interest.